\documentclass[conference]{IEEEtran}
\IEEEoverridecommandlockouts
\usepackage{cite}
\usepackage{amsmath,amssymb,amsfonts}
\usepackage{graphicx}
\usepackage{textcomp}
\usepackage{xcolor}
\def\BibTeX{{\rm B\kern-.05em{\sc i\kern-.025em b}\kern-.08em
    T\kern-.1667em\lower.7ex\hbox{E}\kern-.125emX}}

\usepackage{subcaption}
\usepackage{caption}
\usepackage{enumitem}
\usepackage{amsmath} 
\usepackage{booktabs} 
\usepackage{adjustbox} 
\usepackage{bm}
\usepackage{tabularx} 
\usepackage{graphicx}     
\usepackage{makecell}     
\usepackage{array}        
\usepackage{graphicx}      
\usepackage{algorithm}
\usepackage{bm}
\usepackage{tabularx} 

\usepackage{graphicx}     
\usepackage{booktabs}     

\usepackage{makecell}     
\usepackage{array}        

\usepackage{pifont} 
\usepackage{ragged2e} 

\usepackage{algorithm}
\usepackage{algpseudocode}

\usepackage{pifont}
\newcommand{\xmark}{\ding{55}} 
\newcommand{\cmark}{\checkmark} 

\DeclareRobustCommand{\method}[1]{{\fontsize{7.5}{11}\selectfont \textbf{#1}}}

\begin{document}
\bstctlcite{IEEEexample:BSTcontrol}

\title{DAO-GP Drift Aware Online Non-Linear Regression Gaussian-Process}

\author{\IEEEauthorblockN{1\textsuperscript{st} Mohammad Abu-Shaira}
\IEEEauthorblockA{\textit{Computer Science and Engineering} \\
\textit{The University of North Texas}\\
Denton, USA \\
ShairaAbu-Shaira@my.unt.edu}
\and
\IEEEauthorblockN{2\textsuperscript{nd} Ajita Rattani}
\IEEEauthorblockA{\textit{Computer Science and Engineering} \\
\textit{The University of North Texas}\\
Denton, USA \\
Ajita.Rattani@unt.edu}
\and
\IEEEauthorblockN{3\textsuperscript{rd} Weishi Shi}
\IEEEauthorblockA{\textit{Computer Science and Engineering} \\
\textit{The University of North Texas}\\
Denton, USA \\
Weishi.Shi@unt.edu}
}

\maketitle

\begin{abstract}
Real-world datasets often exhibit temporal dynamics characterized by evolving data distributions. Disregarding this phenomenon, commonly referred to as concept drift, can significantly diminish a model’s predictive accuracy. Furthermore, the presence of hyperparameters in online models exacerbates this issue. These parameters are typically fixed and cannot be dynamically adjusted by the user in response to the evolving data distribution. Gaussian Process (GP) models offer powerful non-parametric regression capabilities with uncertainty quantification, making them ideal for modeling complex data relationships in an online setting. However, conventional online GP methods face several critical limitations, including a lack of drift-awareness, reliance on fixed hyperparameters, vulnerability to data snooping, absence of a principled decay mechanism, and memory inefficiencies. In response, we propose DAO-GP (Drift-Aware Online Gaussian Process), a novel, fully adaptive, hyperparameter-free, decayed, and sparse non-linear regression model. DAO-GP features a built-in drift detection and adaptation mechanism that dynamically adjusts model behavior based on the severity of drift. 
Extensive empirical evaluations confirm DAO-GP’s robustness across stationary conditions, diverse drift types (abrupt, incremental, gradual), and varied data characteristics. Analyses demonstrate its dynamic adaptation, efficient in-memory and decay-based management, and evolving inducing points. Compared with state-of-the-art parametric and non-parametric models, DAO-GP consistently achieves superior or competitive performance, establishing it as a drift-resilient solution for online non-linear regression.
\end{abstract}

\begin{IEEEkeywords}
Online Non-Linear Regression Gaussian Process, Decay, Sparse, Hyperparameter-Free, Concept Drift
\end{IEEEkeywords}

\section{Introduction}
In today's highly dynamic, data-centric environment, marked by a continuous influx of information from diverse sources such as IoT devices, financial markets, and user interactions, the demand for real-time learning and adaptive capabilities has become increasingly critical \cite{hashem2015rise}. However, traditional batch learning falls short due to assumptions of stationary distributions, need for complete datasets, and lack of continuous updates. These inherent drawbacks severely impede their efficacy in dynamic environments. For instance, models trained on historical stock market data often fail to respond to real-time shifts in market sentiment or economic conditions, leading to outdated predictions. Similarly, traffic navigation systems may struggle to efficiently adjust routes when unexpected accidents or congestion occur, causing delays and user frustration. These challenges highlight the growing need for online learning approaches that can immediately adapt to new data, supporting timely, scalable, and adaptive decision-making in dynamic contexts.

Regression models predict continuous outcomes by capturing relationships between variables. While linear regression is interpretable, it struggles with complex, real-world patterns \cite{montgomery2021introduction}. Nonlinear models handle such complexity \cite{seber2003nonlinear} but remain static, making them less effective in evolving data streams. Online regression addresses this by adapting in real time, offering scalability and low latency for dynamic applications like market forecasting \cite{hand2012knowledge}.

While online regression offers a robust approach for adapting to evolving data, its real-world application faces significant hurdles. Concept drift poses a major challenge, as changing data distributions compromise model accuracy \cite{tsymbal2004problem}. The presence of hyperparameters further complicates matters, making manual tuning impractical with frequent concept shifts \cite{liu2023online, lacombe2021meta}. Moreover, growing memory footprints from accumulating data and model components lead to scalability issues \cite{dean2012large}. The lack of a decay mechanism exacerbates these problems, making it difficult to balance adaptability with model stability \cite{loshchilovdecoupled}.

Among online regression models, parametric approaches such as polynomial regression rely on a fixed number of parameters and a predetermined functional form. These models are generally efficient and interpretable but tend to perform poorly when the underlying data relationships diverge from their assumptions \cite{bishop2006pattern}. In contrast, nonparametric models such as Gaussian Processes (GPs) forgo a fixed structure and instead define a distribution over functions \cite{williams2006gaussian}. This allows GPs to flexibly model complex, nonlinear patterns and quantify predictive uncertainty \cite{de2013bayesian}. While this adaptability makes them well-suited for dynamic, evolving data streams, it often comes with higher computational costs \cite{quinonero2005unifying}.

Existing online Gaussian Process (GP) models face significant limitations, primarily their lack of drift awareness, which hampers their ability to adapt to evolving data distributions. These models also typically lack decay mechanisms crucial for down-weighting outdated information in non-stationary environments. Furthermore, their reliance on fixed model hyperparameters and the computational inefficiency of optimizing kernel hyperparameters on every incoming batch misaligns with the nature of concept drift, where adaptation should be driven by detected distributional shifts. Many current online GP models are also kernel-dependent, requiring predefined kernels that introduce data snooping, and some are further constrained by only supporting kernels for structured interpolation, thereby restricting the diversity and flexibility of applicable kernels and limiting their capacity to capture complex data relationships.

In light of these challenges, this work introduces DAO-GP (Drift-Aware Online Gaussian Process), a novel online regression model designed for robust and adaptive learning in dynamic environments. DAO-GP features a built-in in-memory drift detection and adaptation mechanism that not only identifies distributional shifts with a bounded rate of false positives, but also classifies them by magnitude, enabling targeted adaptation strategies. It triggers kernel hyperparameter optimization only upon drift detection, thereby avoiding the inefficiencies of per-batch tuning. The model maintains sparsity through inducing points and integrates a decay mechanism to down-weight outdated data, ensuring responsiveness to recent trends while reducing memory overhead. DAO-GP is inherently hyperparameter-free, an essential property for online learning, where continuous data flow makes manual tuning infeasible. Furthermore, it employs a diverse kernel pool to adaptively respond to distributional changes. Computational efficiency is achieved via the Woodbury Matrix Identity, which reduces the complexity of kernel updates from $O(n^3)$ to $O(m^2n)$, where $m \ll n$, enabling scalable and memory-efficient learning in streaming settings. The source code and datasets utilized can be accessed through our GitHub  repository \cite{DAO-GP-GIT}.

\section{Related Work}
Among parametric approaches, Kernelized Passive-Aggressive Regression (KPA) \cite{crammer2006online} extends the passive-aggressive learning paradigm to nonlinear regression by leveraging kernel functions. The model updates only when the current prediction violates an $\epsilon$-insensitive loss margin, offering computational efficiency and robustness in streaming environments. Kernel Recursive Least Squares (KRLS) \cite{engel2004kernel} generalizes the traditional Recursive Least Squares (RLS) algorithm to nonlinear settings through Mercer kernels, enabling linear regression in a high-dimensional feature space. A key feature is its online sparsification mechanism, which retains only data points that provide significant new information, improving scalability in online scenarios.

\begin{table*}[ht]
\centering
\captionsetup{justification=centering}
\caption{Comparison of Features Across Online Regression Models\\
{\tiny Hparam. : Hyperparameter, Opt.: Optimization, LR: Learning Rate, VI: Variational Inference, $k(\cdot, \cdot)$): kernel function and its parameters}
}
\vspace{-.2cm}
\scriptsize
\setlength{\tabcolsep}{4pt} 
\renewcommand{\arraystretch}{1.1} 
\begin{tabular}{|p{1.7cm}|
>{\centering\arraybackslash}p{1.3cm}|
>{\centering\arraybackslash}p{0.9cm}|
>{\centering\arraybackslash}p{0.9cm}|
>{\centering\arraybackslash}p{3cm}|
>{\centering\arraybackslash}p{2.3cm}|
>{\centering\arraybackslash}p{1.4cm}|
p{4.3cm}|}
\hline
\textbf{Model} &
\textbf{\shortstack{Drift Aware}} &
\textbf{\shortstack{Sparse}} &
\textbf{\shortstack{Decayed}} &
\textbf{\shortstack{Hparam. Free}} &
\textbf{\shortstack{Hparam. Opt.}} &
\textbf{\shortstack{Kernel Indep.}} &
\textbf{\shortstack{Other Limitations}} \\
\hline

\textbf{KPA} \cite{crammer2006online} & \xmark & \cmark & \xmark & \xmark \hspace{1pt} ($C$, $\epsilon$, $k(\cdot, \cdot)$) & \xmark & \xmark &  \\
\hline

\textbf{KRLS} \cite{engel2004kernel} & \xmark & \cmark & \xmark & \xmark \hspace{1pt} ($\lambda$, $k(\cdot, \cdot)$) & \xmark & \xmark &  \\
\hline

\textbf{SOGP} \cite{csato2002sparse} & \xmark & \cmark & \xmark & \xmark \hspace{1pt} ($\epsilon$, $k(\cdot, \cdot)$) & \xmark & \xmark & 
• Sensitivity to $\epsilon$
• Heuristic pruning \\
\hline

\textbf{POG} \cite{koppel2021consistent} & \xmark & \cmark & \xmark & \xmark \hspace{1pt} ($\epsilon$, $k(\cdot, \cdot)$) & \xmark & \xmark & 
• Not pure online 
• Careful tuning needed \\
\hline

\textbf{SVGP} \cite{bui2017streaming} & \xmark & \cmark & \xmark & \xmark \hspace{1pt} (inducing pts., LR, $k(\cdot, \cdot)$) & \cmark (via VI) & \xmark &  \\
\hline

\textbf{OLVGP} \cite{wang2024online} & \xmark & \cmark & \xmark & \xmark \hspace{1pt} (LR, optimizer, $k(\cdot, \cdot)$) & \cmark (per batch) & \xmark & 
• Only structured kernels (e.g., stationary) \\
\hline

\textbf{WISKI} \cite{stanton2021kernel} & \xmark & \cmark & \xmark & \xmark \hspace{1pt} (LR, optimizer, $k(\cdot, \cdot)$) & \cmark (per batch) & \xmark &  \\
\hline

\textbf{OSMGP} \cite{ranganathan2010online} & \xmark & \cmark & \xmark & \xmark \hspace{1pt} ($\epsilon$, $k(\cdot, \cdot)$) & \xmark & \xmark & 
• Basis set may grow unbounded \\
\hline

\textbf{iVSGPR} \cite{cheng2016incremental} & \xmark & \cmark & \xmark & \xmark \hspace{1pt} ($k(\cdot, \cdot)$) & \xmark & \xmark &  \\
\hline

\textbf{FITC-GP} \cite{huber2014recursive} & \xmark & \cmark & \xmark & \xmark \hspace{1pt} ($k(\cdot, \cdot)$) & \cmark (periodic) & \xmark & 
• Requires initial inducing set \\
\hline

\end{tabular}
\label{tab:online-regression-models-comparison}
\end{table*}

In contrast, non-parametric models like Sparse Online Gaussian Process (SOGP) \cite{csato2002sparse} offer efficient online updates by maintaining a sparse set of informative support points. Online FITC-based Gaussian Process models \cite{huber2014recursive,bijl2015online,kania2021fast} extend this idea by maintaining a fixed set of inducing points and updating the posterior recursively using sufficient statistics. These approaches avoid variational inference and do not rely on support set expansion, making them well-suited for memory-efficient online regression. The Projected Online GP (POG)\cite{koppel2021consistent} enables constant-time updates through projection techniques, making it highly scalable for large streams. Streaming SVGP \cite{bui2017streaming} provides a principled variational framework for streaming data using inducing points to approximate the posterior. Online Learning of Variational GP (OLVGP) \cite{wang2024online} supports real-time updates of both inducing variables and hyperparameters using natural gradient and variational inference. WISKI \cite{stanton2021kernel} achieves constant-time inference by leveraging structured kernel interpolation and grid-based inducing structures. OSMGP \cite{ranganathan2010online} offers fully online regression by incrementally updating a sparse GP model using a projection-based basis selection mechanism. Finally, iVSGPR \cite{cheng2016incremental} employs variational inference for probabilistic and efficient online learning while maintaining sparsity in the support set.

The comparative analysis, summarized in Table \ref{tab:online-regression-models-comparison}, reveals that most existing online regression models are limited in their adaptability, particularly regarding drift awareness, decay mechanisms, and hyperparameter independence. While models like KRLS, SOGP, and Streaming SVGP achieve sparsity, they lack inherent strategies for managing concept drift or discarding outdated data. Furthermore, none of the surveyed models are truly hyperparameter-free, requiring manual tuning of thresholds, learning rates, or kernel parameters, which often makes them less practical for true streaming environments characterized by drift. Although variational approaches do support online hyperparameter optimization, they perform this optimization on every incoming batch, proving inefficient compared to triggering it only when drift is detected. These findings collectively underscore the critical need for a fully adaptive framework to overcome these pervasive limitations.

\section{Method}

\subsection{Problem Settings}

Consider $\mathcal{X} \subseteq \mathbb{R}^d$ as the input feature space and $\mathcal{Y} \subseteq \mathbb{R}$ as the continuous output space. At each time step $t$, a new data point $(\mathbf{x}_t, y_t)$ is received, where $\mathbf{x}_t \in \mathcal{X}$ is the input vector and $y_t \in \mathcal{Y}$ is the corresponding real-valued target. The task is to learn the underlying relationship $y_t = f(\mathbf{x}_t) + \epsilon_t$, where $f$ is an unknown, possibly complex non-linear function and $\epsilon_t$ denotes Gaussian noise, $\epsilon_t \sim \mathcal{N}(0, \sigma_n^2)$.

In our approach, the latent function $f$ is modeled as a random function drawn from a Gaussian Process (GP) prior. A GP is fully specified by its mean function $m(\mathbf{x}) = \mathbb{E}[f(\mathbf{x})]$ and covariance function (kernel) $k(\mathbf{x}, \mathbf{x}') = \mathbb{E}[(f(\mathbf{x}) - m(\mathbf{x}))(f(\mathbf{x}') - m(\mathbf{x}'))]$. For simplicity, the mean function is typically assumed to be zero, i.e., $m(\mathbf{x}) = 0$. The GP prior is denoted as $f(\mathbf{x}) \sim \mathcal{GP}(m(\mathbf{x}), k(\mathbf{x}, \mathbf{x}'))$, leading to observations of the form $y_t = f(\mathbf{x}_t) + \epsilon_t$.

The data stream is assumed to be potentially infinite, requiring models that can update sequentially with limited memory resources \cite{hoi2021online}. Additionally, real-world data streams often exhibit concept drift, where the underlying data distribution evolves over time. Formally, given a stream $\{(\mathbf{x}_1, y_1), (\mathbf{x}_2, y_2), \dots\}$ generated from a joint distribution $p(\mathbf{x}, y)$, concept drift occurs if:
$\exists \mathbf{x}, y \; : \; p_{t_n}(\mathbf{x}, y) \neq p_{t_{n+1}}(\mathbf{x}, y)$,
where $t_n$ and $t_{n+1}$ are consecutive time points. This drift can be abrupt, gradual, or incremental, each posing unique challenges to the learning system \cite{webb2016characterizing,gama2014survey}.

The objective of the proposed online non-linear regression model is to incrementally update the predictive function $f_t: \mathcal{X} \rightarrow \mathcal{Y}$ by optimizing the GP marginal likelihood. This is equivalent to minimizing the negative log-marginal likelihood (NLML) at each step. Given the data observed up to time $t$, $\mathcal{D}_t = \{(\mathbf{x}_i, y_i)\}_{i=1}^t$, the NLML is defined as \cite{murphy2012machine}:
\[
\mathcal{L}(\boldsymbol{\theta}) = \frac{1}{2} \mathbf{y}_t^\top (\mathbf{K}_t + \sigma_n^2 \mathbf{I})^{-1} \mathbf{y}_t + \frac{1}{2} \log |\mathbf{K}_t + \sigma_n^2 \mathbf{I}| + \frac{t}{2} \log(2\pi),
\]
where $\mathbf{K}_t$ is the kernel matrix with $K_{ij} = k(\mathbf{x}_i, \mathbf{x}_j)$, $\sigma_n^2$ is the noise variance, and $\boldsymbol{\theta}$ are the kernel hyperparameters. This formulation provides a principled trade-off between data fit and model complexity.

\subsection{DAO-GP Architecture and Flow Process}

\begin{figure}[htbp]
  \centering
  \includegraphics[width=0.45\textwidth, height=4.5cm]{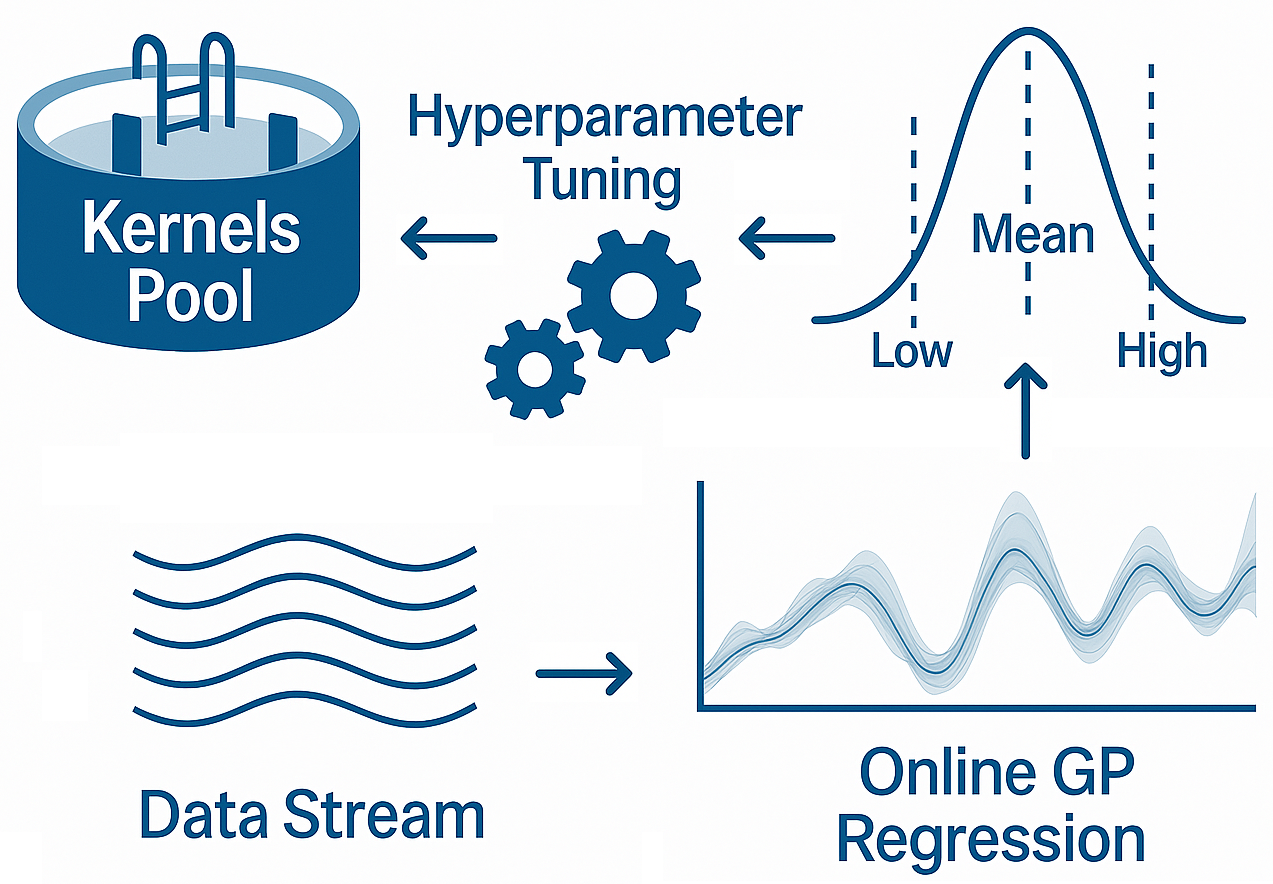}
  \caption{DAO-GP Architecture and Flow Process}
  \label{fig:DAO-GP-Components-And-WorkFlow}
  \vspace{-1.2cm}
\end{figure}

The DAO-GP framework is engineered for superior performance in non-stationary data streams, seamlessly integrating four critical components to ensure robust adaptability. At its core, the \textbf{Online Gaussian Process (GP) Regression Engine} continuously updates its predictive function with each new data instance, with adaptation intrinsically tied to real-time concept drift assessment. This is complemented by a \textbf{Memory-Based Online Drift Detection and Magnitude Quantification} mechanism, which monitors and records performance indicators (KPIs) of incoming data, proactively identifying concept drift and categorizing its severity. A sophisticated \textbf{Hyperparameters Optimization} component dynamically refines the GP’s predictive accuracy by optimizing hyperparameters using the Negative Log Marginal Likelihood (NLML). Finally, a \textbf{Dynamic Kernel Pool for Adaptive Selection} enables DAO-GP to select the most appropriate kernel from a predefined set, ensuring responsiveness and robust adaptation to even the most severe distributional shifts. The synergistic interplay of these components is illustrated in Figure~\ref{fig:DAO-GP-Components-And-WorkFlow}.
The DAO-GP framework is meticulously designed as a continuous, adaptive pipeline for effectively managing non-stationary data streams. Its operation commences with the real-time ingestion of incoming data instances, processed either individually or in mini-batches. Each new observation then triggers an incremental update to the online Gaussian Process (GP) regression model, dynamically allowing its predictive function to continuously adapt to the evolving data landscape. Throughout this ongoing process, critical performance KPIs, specifically Mean Squared Error (MSE) and R-squared (R$^2$), are rigorously monitored to provide immediate insights into the model's predictive accuracy and goodness-of-fit. A paramount subsequent step involves sophisticated in-memory drift detection and categorization, where the system quantifies any observed distributional shifts within the data, classifying them with granular detail as either a subtle incremental change or a pronounced abrupt shift, thereby enabling targeted adaptive responses.

The framework's adaptive response is intricately contingent upon the precise categorization of detected drift. Upon the identification of a minor (incremental) drift, DAO-GP initiates a focused hyperparameter optimization routine. This process minimizes the Negative Log-Marginal Likelihood (NLML) using a buffer of recent observations, fine-tuning the model's internal parameters to accommodate gradual distributional shifts. Conversely, the detection of a major (abrupt) drift triggers a more profound and multi-layered adaptation strategy. This begins with an initial round of hyperparameter optimization (also via NLML), followed by re-evaluation of the drift's magnitude. If the significant drift persists, the system activates a dynamic kernel selection mechanism, judiciously choosing the most appropriate kernel from a predefined pool. This re-alignment of the model with new data characteristics ensures optimal performance under severe distributional changes. This systematic and dynamically responsive workflow, combined with its modular architecture, empowers DAO-GP to proportionately address varying degrees of non-stationarity. The approach strikes a critical balance between adaptive flexibility and predictive robustness, sustaining high performance across diverse and complex data stream conditions. The algorithmic steps of DAO-GP are shown in Algorithm~\ref{alg:dao-gp-alg}.

\begin{algorithm}
\caption{DAO-GP Drift-Aware Online Gaussian-Process}
\label{alg:dao-gp-alg}
\footnotesize
\begin{algorithmic}[1]

\State \textbf{Input:} Data stream \(\{ \mathbf{X}, \mathbf{y} \}\), 
  \textit{max-inducing}, \(\gamma\), \textit{initial-kernel}, \textit{ik-threshold},\\\hspace{23pt}\textit{kernel-pool} $\subseteq$ \(\{\text{Rbf, Poly, etc.\}}\) , \textit{uncertainty-threshold}, $\zeta$ \\\hspace{21pt}\textit{initial-batch-size}, \( \rho \in (0,1] \), \( \textit{kpi} \subseteq \{\text{Acc, Loss, etc.}\}\)
  
\State Initialize an empty list: $L \gets [\hspace{2pt}]$ \makebox[0pt][l]{\hspace{1.8cm}\Comment{\scriptsize $L$ is the KPI window}}

\State $\mathbf{X}_\text{base-tr}$, $\mathbf{y}_\text{base-tr}$, $\mathbf{X}_\text{base-vl}$, $\mathbf{y}_\text{base-tr}$ = \method{MBTrainValSplit}({$\mathbf{X}_\text{base}$, $\mathbf{y}_\text{base}$})

\State  $\textit{kernel}$, $\mathbf{K}_{\text{fn}}$, $\mathbf{K}$, $\mathbf{K^{-1}}$, $\mathbf{K}_{\text{OPT-Hprm}}$, $\mathbf{\mu_*}$, $\mathbf{\sigma^2_*}$, $\mathbf{MB}_{\text{kpi}}$ $\gets$ \method{PickBestKernel}($\mathbf{X}_{\text{base-tr}}$, $\mathbf{y}_{\text{base-tr}}$, 
$\mathbf{X}_{\text{base-vl}}$, $\mathbf{y}_{\text{base-vl}}$, \textit{initial-kernel}, \textit{ik-threshold}, \\\hspace{13pt}\textit{input-dim}, \textit{kernel-pool}, \textit{kpi})


\State \method{AddItem}($L$, $\mathbf{MB}_{\text{kpi}}$)

\For{$t \gets 1$ to $T$}
    \State $\mathbf{X}_\text{inc-tr}$, $\mathbf{y}_\text{inc-tr}$, $\mathbf{X}_\text{inc-vl}$, $\mathbf{y}_\text{inc-tr}$ = \method{MBTrainValSplit}({$\mathbf{X}_\text{inc}$, $\mathbf{y}_\text{inc}$})
    
\ForAll{$(\mathbf{x}_{i}, \mathbf{y}_{i}) \in (\mathbf{X}_{\text{inc-tr}},\,\mathbf{y}_{\text{inc-tr}})$}
        \State $\sigma^2_* \gets$ \method{ComputePredictiveVariance}($\mathbf{X}_\text{inc-tr}$, $\mathbf{y}_\text{inc-tr}$, $\mathbf{x}_{\text{i}}$, $\mathbf{K}_\text{fn}$, \\\hspace{140pt}$\mathbf{K}_{\text{OPT-Hprm}}$)

        \If{$\sigma^2_* > \textit{uncertainty-threshold}$}
            \State $\mathbf{K^{-1}}$ = \method{WoodburySafeUpdate}($\mathbf{K^{-1}}$, $\mathbf{X}_\text{base-trn}$, $\mathbf{x}_{\text{i}}$, $\mathbf{K}_\text{fn}$, \\\hspace{138pt}$\mathbf{K}_{\text{OPT-Hprm}}$)
                
                \State $\mathbf{X}_{\text{base-tr}} \gets \begin{bmatrix} \mathbf{X}_{\text{base-tr}} \\ \mathbf{x}_{\text{i}} \end{bmatrix}$
                \State $\mathbf{y}_{\text{base-tr}} \gets \mathbf{y}_{\text{base-tr}} \cup \mathbf{y}_{\text{i}}$                
        \EndIf
    \EndFor
    \\

    \State \scriptsize \texttt{/* Start kernel hyperparameter optimization once the}\\\hspace{18pt} \texttt{ KPI window reaches its threshold size. */}

    \State $\mu_{*}, \sigma^2_{*}, \mathbf{K^{-1}} \gets $ \method{ComputeGP}($\mathbf{X}_{\text{base-tr}}$, $\mathbf{y}_{\text{base-tr}}$, $\mathbf{X}_\text{inc-vl}$, $\mathbf{K}_\text{fn}$, $\mathbf{K}_{\text{OPT-Hprm}}$)

    \State $\mathbf{MB}_{\text{kpi}}  \gets$ \method{MiniBatchKPIs}($\mathbf{X}_{\text{base-tr}}$, $\mathbf{y}_{\text{base-tr}}$, $\mathbf{X}_{\text{inc-vl}}$, $\mu_{*}$)
    
     \State $ \tau, \mu, \sigma, \text{low}, \text{high}, \text{DM} \gets \method{MeasureKPIs}(L, z)$\Comment{\footnotesize $z \gets \Phi^{-1}(1 - \rho)$}
     
    \State $\textit{drift} \gets \method{DetectEvaluateDrift}(\mu, L_{t\textit{-}1}, \tau, \zeta)$ 
    \If{\textit{drift}}  
        \State \method{RemoveItem}($L$, $L_{t\text{-}1}$)            

        \State $\mathbf{K}_{\text{OPT-Hprm}} \gets$ \method{OptimizeHparams}($\mathbf{X}_{\text{base-tr}}, \mathbf{y}_{\text{base-tr}}$, \textit{kernel}, \\\hspace{135pt}$\mathbf{K}_{\text{OPT-Hprm}}$)

        \State $\mu_{*}, \sigma^2_{*}, \mathbf{K^{-1}} \gets $ \method{ComputeGP}($\mathbf{X}_{\text{base-tr}}$, $\mathbf{y}_{\text{base-tr}}$,  $\mathbf{X}_\text{inc-vl}$, $\mathbf{K}_\text{fn}$, \\\hspace{125pt}$\mathbf{K}_{\text{OPT-Hprm}}$)

        \State $\mathbf{MB}_{\text{kpi}}  \gets$ \method{MiniBatchKPIs}($\mathbf{X}_{\text{base-tr}}$, $\mathbf{y}_{\text{base-tr}}$, $\mathbf{X}_{\text{inc-vl}}$, $\mu_{*}$)
        
        \State \method{AddItem}($L$, $\mathbf{MB}_{\text{kpi}}$)

        \State $\textit{drift, drift-type} \gets \method{DetectEvaluateDrift}(\mu, L_{t\text{-}1}, \tau, \zeta)$

        \If{\textit{drift-type} = \texttt{Abrupt}}
            \State \method{RemoveItem}($L$, $L_{t\text{-}1}$)            
            \State \textit{kernel}, $\mathbf{K}_{\text{fn}}$, $\mathbf{K}_{\text{OPT-Hprm}}$, $\mathbf{MB}_{\text{kpi}} \gets$ \method{PickBestKernel}($\mathbf{X}_{\text{base-tr}}$, \\\hspace{100pt}$\mathbf{y}_{\text{base-tr}}$,
            $\mathbf{X}_{\text{inc-vl}}$, $\mathbf{y}_{\text{inc-vl}}$, \textit{kernel}, \\\hspace{100pt}\textit{ik-threshold}, \textit{input-dim}, \textit{kernel-pool}, \textit{kpi})

            \State \method{AddItem}($L$, $\mathbf{MB}_{\text{kpi}}$)        
        \EndIf
    \EndIf

    \State $\mathbf{X}_{\text{base-tr}}$, $\mathbf{y}_{\text{base-tr}}$, $\mathbf{K^{-1}}$ = \method{SelectInducingPoints}($\mathbf{X}_{\text{base-tr}}$, $\mathbf{y}_{\text{base-tr}}$, \\\hspace{100pt}\textit{max-inducing}, $\gamma$, $\mathbf{t}_{\text{base-trn}}$, $\mathbf{K}_{\text{fn}}$, 
    \\\hspace{100pt}$\mathbf{K}_{\text{OPT-Hprm}}$, $\mathbf{K}$, $\mathbf{K^\text{-1}}$)

    \State $\mu_{*}, \sigma^2_{*} \gets $ \method{ComputeGP}($\mathbf{X}_{\text{base-tr}}$, $\mathbf{y}_{\text{base-tr}}$, $\mathbf{K}_\text{fn}$,  $\mathbf{K}_{\text{OPT-Hprm}}$)

\EndFor

\State \textbf{return} $\mu_{*}, \sigma^2_{*}$, $\mathbf{K}$, $\mathbf{K^{-1}}$, \textit{kernel}

\end{algorithmic}
\end{algorithm}

\begin{algorithm}
\caption{\method{SelectInducingPoints}}
\label{alg:select-inducing-points}
\begin{algorithmic}[1]

\State \textbf{Input:} $\mathbf{X}_\text{tr}$, $\mathbf{y}_\text{tr}$, \textit{max-inducing}, $\gamma$, 
$\mathbf{t}_{\text{base-trn}}$, $\mathbf{K}_{\text{fn}}$, \\\hspace{27pt}$\mathbf{K}_{\text{OPT-Hprm}}$, $\mathbf{K}$, $\mathbf{K^\text{-1}}$

\If{$|\mathbf{X}_{\text{trn}}| \leq \textit{max-inducing}$}
    \State return $\mathbf{X}_\text{tr}$, $\mathbf{y}_\text{tr}$, $\mathbf{K^\text{-1}}$
\EndIf

\State $\mathbf{w} \gets \method{ComputeDecayWeights}(\gamma, \mathbf{t}_{\text{base-trn}})$

\State $\mathbf{K}_{\text{decayed}}[i,j] \gets \sqrt{w_i} \cdot \mathbf{K}[i,j] \cdot \sqrt{w_j}, \quad \forall i, j$

\State $\sigma^2_{*} \gets$ \method{ComputePredictiveVariance}($\mathbf{K}_{\text{decayed}}$) 

\State $\text{scores} \gets \mathbf{w} \odot \boldsymbol{\sigma}^2_*$

\State $\mathcal{I}$ $\gets$ \method{TopScores}(scores, \textit{max-inducing})

\State $\mathbf{X}_\text{tr}$, $\mathbf{y}_\text{tr}$, $\mathbf{t}_{\text{base-trn}}$ $\gets$ $\mathbf{X}_{\text{tr}}[\mathcal{I}]$, $\mathbf{y}_{\text{tr}}[\mathcal{I}]$, $\mathbf{t}_{\text{base-trn}}[\mathcal{I}]$

\State $\mathbf{K}$, $\mathbf{K^{-1}}$ $\gets$ $\mathbf{K}_{\text{fn}}$($\mathbf{X}_\text{tr}$, $\mathbf{y}_\text{tr}$, $\mathbf{K}_{\text{OPT-Hprm}}$)

\State \textbf{return} $\mathbf{X}_\text{tr}$, $\mathbf{y}_\text{tr}$, $\mathbf{K^{-1}}$

\end{algorithmic}
\end{algorithm}
\vspace{-.3cm}

\subsection{Online Learning Parameter Update}
At each discrete time step, the proposed DAO-GP model incrementally updates its internal parameters in response to newly received data, whether as single instances or mini-batches. The process begins with an initial mini-batch ($\mathbf{X}_\text{base}$, $\mathbf{y}_\text{base}$) used for model initialization, the initial batch size is determined using a configuraton parameter (\textit{initial-batch-size}). The optimal kernel function is then selected from a predefined pool based on validation-driven performance metrics, and its corresponding kernel matrix inverse is computed. Notably, the framework supports user-specified kernel enforcement: if a kernel is explicitly provided by the user, it is retained as long as its performance remains within a predefined tolerance (\textit{ik-threshold}) of the best-performing kernel identified during validation.

In the kernel selection phase, each candidate kernel $k_i \in \mathcal{K}$ is evaluated by training a Gaussian Process on the base training set $(\mathbf{X}_{\text{base-tr}}, \mathbf{y}_{\text{base-tr}})$ and computing its posterior prediction on a held-out validation set $\mathbf{X}_{\text{base-vl}}$. Let $K$ denote the kernel matrix over $\mathbf{X}_{\text{base-tr}}$, $K_s$ the cross-kernel between training and validation inputs, and $K_{ss}$ the kernel matrix over validation inputs. The predictive mean and variance are given by Equations \ref{eq:eq_mean_base} and \ref{eq:eq_variance_base} \cite{williams1998prediction,seeger2004gaussian}, respectively.
\begin{equation}
\label{eq:eq_mean_base}
\boldsymbol{\mu}_* = K_s^\top (K + \sigma^2 I)^{-1} \mathbf{y}_{\text{base-tr}}
\end{equation}
\begin{equation}
\label{eq:eq_variance_base}
\boldsymbol{\sigma}^2_* = \mathrm{diag} \left( K_{ss} - K_s^\top (K + \sigma^2 I)^{-1} K_s \right)
\end{equation}

The predictive mean is used to evaluate validation performance based on a user-specified \textit{kpi} (e.g., R\textsuperscript{2} or MSE), and the kernel that achieves the best performance is selected. This step returns the selected kernel type (e.g. rbf), the kernel matrix \( K \), its inverse \( K^{-1} \), the corresponding kernel function, the optimized hyperparameters, and the predictive outputs, including the mean \( \mu_* \) and variance \( \sigma^2_* \). These components are then used to initialize the base model and support subsequent online updates.

At each time step $t$, a new data point or mini-batch $(\mathbf{X}_\text{inc}, \mathbf{y}_\text{inc})$ is received and partitioned into training and validation subsets. In cases where the mini-batch contains only a single instance, the training and validation sets are identical. For \textit{each instance} $(\mathbf{x}_i, y_i)$ in the training portion $(\mathbf{X}_\text{inc}, \mathbf{y}_\text{inc})$, the model computes the predictive variance $\sigma^2_*$ at $\mathbf{x}_i$ given the current model using the current kernel inverse. If the variance $\sigma^2_*$ exceeds a predefined \textit{uncertainty threshold}, indicating that the new data point provides significant new information or falls into an uncertain region, the inverse kernel matrix \( \mathbf{K}^{-1} \) is updated using a rank-one form of the \textit{Woodbury matrix identity}\cite{henderson1981deriving}, and the new informative point is added to the base training set $(\mathbf{X}_{\text{base-tr}}, \mathbf{y}_{\text{base-tr}})$. The updated training set and inverse kernel are then used to make predictions ($\mu_*$) for the current mini-batch $(\mathbf{X}_\text{inc}, \mathbf{y}_\text{inc})$.

\begin{equation}
\mathbf{K}_{t+1}^{-1} = 
\begin{bmatrix}
\mathbf{K}_t^{-1} + \beta \mathbf{a} \mathbf{a}^\top & -\beta \mathbf{a} \\
-\beta \mathbf{a}^\top & \beta
\end{bmatrix},
\label{eq:woodbury}
\end{equation}

In this formulation of the Woodbury identity matrix, identified by Equation~\ref{eq:woodbury}, \( \mathbf{a} = \mathbf{K}_t^{-1} \mathbf{k} \), where \( \mathbf{k} \) is the kernel vector between the current base inputs and the new data point. The scalar \( c \) is computed as \( c = k(\mathbf{x}_i, \mathbf{x}_i) + \sigma^2 \), and the scaling coefficient is defined as \( \beta = \left( c - \mathbf{k}^\top \mathbf{K}_t^{-1} \mathbf{k} \right)^{-1} \). This update provides an efficient way to expand the inverse kernel matrix without recomputing the full inverse from scratch. However, if the denominator becomes too small, which indicates potential numerical instability, the update is discarded and the inverse of the current kernel matrix is recomputed from scratch.

This initial model update is followed by concept drift detection, categorization, and adaptation, either via memory-based optimization or kernel pool selection in the case of persistent abrupt drift. To control memory usage, inducing point selection and decay mechanisms are subsequently applied, as detailed in the following subsections.

\subsection{Memory-Based Online Drift Handling and Hyperparameter Tuning}

DAO-GP features a built-in hyperparameter drift handling mechanism to categorize minor drifts (incremental) and abrupt (sudden) drifts. This approach leverages historical Key Performance Indicators (KPIs), such as MSE and R$^2$, etc., by organizing them into a normal distribution. Through this process, `low' and `high' limits are determined to evaluate performance deviations of the new decision boundary and quantify the drift. Subsequently, if drift is identified, the kernel's hyperparameters are optimized. If, even after this optimization, the drift persists and is deemed severe, the model is then referred to the predefined kernel pool to facilitate the selection of a more appropriate kernel.

The drift-handling process involves recording performance KPIs at each time \(t\) in a bounded KPI-Window (KPI-Win) of size \textit{KWS}, defined in Equation~\ref{eq:window-size} using recent \(N\) points in the rolling window and mini-batch size \(K\). The scaling factor \(\delta = 0.05\) adjusts \textit{KWS}, constrained between lower (LB) and upper (UB) bounds. An experimental size of 31 is used, though smaller values suit frequent abrupt drift.
\begin{equation}
\label{eq:window-size}
\text{LB} \leq \text{KWS} = \left(\frac{N}{K}\right) \times \delta \leq \text{UB}    
\end{equation}

In an illustrative example with a KPI-Win of size 31, the first 30 entries correspond to the historical \textit{Baseline Statistics}, while the 31st entry represents the KPIs computed using the current mini-batch ($\text{Inst}_{\scriptscriptstyle\text{KPI}}$). The \textit{Baseline Statistics} are normally distributed, facilitating computation of a variable threshold \( \tau \). This eliminates the need for a fixed threshold, which may not suit all KPIs or data streams due to varying characteristics and dynamics.

The threshold is computed using equation~\ref{eq:threshold}, where \( \sigma \) is the standard deviation and \( z \) is the multiplier, indicating number of standard deviations from the mean. The value of $z$ is determined from the user-specified false-alarm probability $\rho$ through the inverse complementary normal cumulative distribution function (CDF) \cite{lages2024hierarchical}, expressed as $z = \Phi^{-1}(1-\rho)$. Here, $\rho$ is the user-defined bound on the false-positive rate, representing the maximum tolerated probability that a benign observation is incorrectly flagged as drift under stable conditions. A smaller $\rho$ produces a larger $z$, resulting in a more relaxed threshold that avoids classifying minor fluctuations as drift, while a larger $\rho$ yields a smaller $z$ and a more sensitive threshold that flags subtle changes as drift.
\vspace{-.2cm}
\begin{equation}
\vspace{-.2cm}
\label{eq:threshold}
\text{Threshold } (\tau) = z \times \sigma
\end{equation}
\begin{equation}
\vspace{-.2cm}
    \label{eq:low}
\text{low}  = \text{\Large $\mu$}_{\scriptscriptstyle\text{KPI}} - \tau  
\end{equation}
\begin{equation}
\vspace{-.2cm}
    \label{eq:high}
     \text{high} = \text{\Large $\mu$}_{\scriptscriptstyle\text{KPI}}+ \tau 
\end{equation}
\begin{equation}            
\label{eq:drift_magnitude_lstm_sccm}
    \text{DM} = \left| \text{\Large $\mu$}_{\scriptscriptstyle\text{KPI}} - \text{ Inst}_{\scriptscriptstyle\text{KPI}} \right|
\end{equation}
\begin{figure}[ht]
\vspace{-.7cm}
  \centering
  \includegraphics[width=0.4\textwidth, height=.15\textheight]{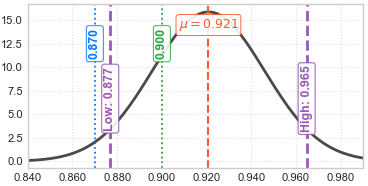}
  \vspace{-.1cm}
  \caption{Limits - Gauss Dist. of KPI-Win on KPI = R$^2$}
  \label{fig:KPI-Normal-Distribution}
  \vspace{-.3cm}
\end{figure}

Figure~\ref{fig:KPI-Normal-Distribution} illustrates the classification of Intant Feeds (Inst) observations relative to a reference performance distribution, using \textit{kpi} = R$^2$ as the evaluation metric. The green line denotes an incoming observation that falls between the mean minus the user-defined \textit{safe area threshold} ($\zeta$) and the lower control limit (Equation~\ref{eq:low}), depicted in the purple line with R$^2$ = 0.877, indicating a minor performance drift. The parameter $\zeta$ defines the \textit{safe area threshold}, which determines whether a performance variation is classified as drift. Specifically, for KPIs that favor higher values (e.g., R$^2$), observations within the range $[\text{mean} - \zeta, \text{mean}]$ are not considered drift. Conversely, for KPIs that favor lower values (e.g., MSE), the safe range is defined as $[\text{mean}, \text{mean} + \zeta]$.

In such cases, exemplified by the instant feed shown by the green lin, the model responds by re-optimizing the kernel hyperparameters using the Negative Log-Marginal Likelihood (NLML) objective. In contrast, the blue line represents an observation that lies beyond the lower control limit, signifying an abrupt performance drift. This severe deviation is initially addressed through NLML-based re-optimization in an attempt to restore acceptable performance. After re-optimization, the KPI is re-evaluated; if the drift remains substantial, the model proceeds to select an alternative kernel from the predefined kernel pool. It is also important to note that the `high' control limit (Equation \ref{eq:high}) is applicable to KPIs where lower values are preferred (e.g., MSE), while the `low' limit is used for KPIs where higher values are favorable (e.g., R$^2$).

In this work, we distinguish between \textit{hyperparameters} and \textit{configuration parameters}. \textit{Hyperparameters} are trainable settings that require optimization (e.g., learning rate, regularization), whereas \textit{configuration parameters} are user-defined choices fixed at deployment (e.g., mini-batch size, KPI). This distinction underpins our claim that DAO-GP is tuning-free (hyperparameter-free), as it relies only on configuration parameters that remain constant during operation.

\subsection{Kernels Optimization \& Kernel Pool}
DAO-GP defines a structured metadata system for kernel configuration through a set of modular classes that specify the hyperparameters for each supported kernel type. Each kernel class provides a clear definition of its tunable parameters, including initial values and valid bounds, enabling systematic optimization and validation. In addition, DAO-GP utilizes Automatic Relevance Determination (ARD) \cite{seeger2004gaussian,tipping2001sparse} kernels, which allow kernel hyperparameters such as length scales to be optimized independently for each input dimension. In such cases, the number of input dimensions is passed to the kernel class to dynamically construct a parameter list that reflects the dimensional structure of the data. This design supports fine-grained control over kernel behavior and enhances adaptability to high-dimensional, heterogeneous data streams.

The optimization mechanism in DAO-GP involves minimizing the NLML, which requires optimizing kernel hyperparameters such as length-scales, variances, and noise levels. L-BFGS-B \cite{byrd1995limited,nocedal2006numerical} is an ideal choice for this task as it is hyperparameter-free, helping avoid additional tuning overhead. This aligns with our vision of an ideal online learning model that prioritizes adaptability and automation over intervention. L-BFGS-B works by iteratively updating parameter estimates using gradient information and a limited-memory approximation of the inverse Hessian matrix \cite{nocedal1980updating}. This approach enables efficient convergence without the need for specified learning rates. In addition, it supports bounded hyperparameter optimization, where parameters are restricted to predefined ranges to maintain numerical stability and avoid degenerate solutions. Its memory efficiency and effectiveness in high-dimensional spaces further support the lightweight and adaptive characteristics that DAO-GP aims to achieve.

\subsection{Inducing Points Selection and Decay Mechanism}

DAO-GP introduces a scoring mechanism for selecting inducing points in streaming environments, detailed in Algorithm~\ref{alg:select-inducing-points}, balances the need for recency, informativeness, and computational tractability. This approach is well-suited for dynamic and large-scale non-stationary data streams.

When the number of candidate training points exceeds the allocated inducing points, DAO-GP employs an exponential decay weighting scheme, defined by $\gamma$. This scheme reduces the influence of older observations based on their timestamps, with each decay weight calculated as $w_i = \exp(-\gamma \cdot \Delta t_i)$. This temporal weighting is integrated into the kernel matrix via a symmetric transformation, where each entry is adjusted as $K_{\text{decayed}}[i,j] = \sqrt{w_i} \cdot K[i,j] \cdot \sqrt{w_j}$. This results in a decay-adjusted kernel that embeds temporal relevance.

To determine the utility of each data point, DAO-GP calculates the predictive variance, $\sigma^2_{,i}$, using this decayed kernel. This variance reflects the model's uncertainty for each instance. These variances are then element-wise multiplied by the decay weights to generate a final score for each point, given by $\text{score}_i = w_i \cdot \sigma^2_{,i}$. This scoring mechanism effectively captures both the uncertainty and recency of each data point. The points with the highest scores are subsequently chosen as inducing inputs. Unlike methods relying on fixed thresholds or heuristic pruning, DAO-GP's scoring strategy adaptively prioritizes recent, high-uncertainty data, making it ideal for dynamic and non-stationary data streams. 

Considering DAO-GP’s theoretical guarantees, we follow the standard online learning assumption that each incoming instance or mini-batch is generated under a stable concept, i.e., no drift occurs within a single feed \cite{hoi2021online,gama2014survey}. Within this setting, DAO-GP builds upon well-established theoretical results: NLML minimization yields consistent hyperparameter estimates in Gaussian Process models under stationary assumptions \cite{williams2006gaussian,seeger2004gaussian}.

The maintained bounded set of inducing points, along with the Woodbury identity update, keeps the time complexity of DAO-GP at $O(M^2)$ per step, ensuring efficient operation regardless of the total data seen. The drift mechanism introduces a lightweight cost of $O(k \cdot d)$ per mini-batch for computing KPI metrics and a higher cost of $O(T \cdot h \cdot n_b^2 + n_b^3)$ only upon drift, due to NLML-based hyperparameter optimization. Here, $M$ is the number of inducing points, $d$ is the input dimension, $k$ is the number of KPI metrics, $T$ is the number of optimizer iterations, $h$ is the number of kernel hyperparameters, and $n_b$ is the mini-batch size. Since drift events are infrequent, this cost is amortized, and the overall complexity remains dominated by the efficient sparse GP updates, preserving DAO-GP’s low per-step computational overhead. In addition, DAO-GP maintains a fixed-size inducing set, leading to a memory cost of $O(M^2 + M \cdot d)$ for storing the kernel matrix and its inverse. Kernel pool parameters and KPI statistics add a small overhead of $O(K \cdot h + |L|)$. Overall, DAO-GP achieves constant memory usage over time, independent of the data stream length.

\section{Experiments}
\begin{table}[t]
\scriptsize
\centering
\captionsetup{justification=centering}
\setlength{\tabcolsep}{3pt}
\caption{Experiment \ref{sec:robustness-under-varying-conditions} – Robustness\\ {\tiny Decay Gamma ($\gamma$ = 0.99), Uncertainty Threshold = 0.001, Initial Kernel = RBF, KPI = R$^2$, $\rho$ = 0.006,  Safe Area Threshold ($\zeta$)= 0.005. Evaluation uses an 80/20 temporal split with no future leakage; datasets are parametric functions.}}
\vspace{-.13cm}
\label{tab:nnr-dataset-full}

\resizebox{\columnwidth}{!}{%
\begin{tabular}{*{5}{c} *{3}{c} *{2}{c}} 
\toprule
\multicolumn{5}{c}{\textbf{Dataset Properties}} 
& \multicolumn{3}{c}{\textbf{Conf. Parameters}} 
& \multicolumn{2}{c}{\textbf{Results}} \\
\cmidrule(lr){1-5} \cmidrule(lr){6-8} \cmidrule(lr){9-10}
\makecell{Datasets} & \makecell{Data\\Points} & \makecell{Dimen-\\sions} & Noise & Focus
& \makecell{Initial\\Batch} & \makecell{Increment\\Size} & \makecell{Max\\Inducing}
& MSE & R$^2$ \\
\midrule
DS01 & 1K & 3 & 10 & Sinusoidal & 20 & 20 & 100 & 0.0419 & 0.9756 \\
DS02 & 10K & 100 & 20 & Sinusoidal & 50 & 50 & 200 & 0.0241 & 0.9753 \\
DS03 & 1K & 3 & 10 & Quadratic & 20 & 20 & 100 & 0.0002 & 0.9998 \\
DS04 & 10K & 100 & 20 & Quadratic & 50 & 50 & 200 & 0.0241 & 0.9753 \\
DS05 & 1K & 3 & 10 & Cubic & 20 & 20 & 100 & 0.0002 & 0.9998 \\
DS06 & 10K & 100 & 20 & Cubic & 50 & 50 & 200 & 0.0440 & 0.9952 \\
DS07 & 1K & 3 & 10 & Exponential & 20 & 2 & 100 & 0.0833 & 0.9796 \\
DS08 & 10K & 100 & 20 & Exponential & 50 & 2 & 200 & 0.0468 & 0.9875 \\
DS09 & 1K & 2 & 10 & Logarithmic & 20 & 1 & 50 & 0.0093 & 0.9785 \\
DS10 & 10K & 200 & 20 & Logarithmic & 30 & 1 & 100 & 0.0402 & 0.9238 \\
DS11 & 0.5K & 5 & 10 & Piecewise & 5 & 5 & 50 & 0.0087 & 0.9928 \\
DS12 & 20K & 200 & 20 & Piecewise & 50 & 50 & 100 & 0.0408 & 0.9694 \\
DS13 & 2K & 3 & 10 & Parabolic Wave & 100 & 4 & 50 & 0.0162 & 0.9902 \\
DS14 & 10K & 300 & 20 & Parabolic Wave & 250 & 250 & 500 & 3.8826 & 0.9603 \\
DS15 & 2K & 100 & 10 & Gaussian & 100 & 20 & 100 & 0.4475 & 0.9979 \\
DS16 & 10K & 20 & 20 & Gaussian & 50 & 1 & 200 & 0.0444 & 0.9989 \\
DS17 & 2K & 10 & 10 & Double Gaussian & 100 & 25 & 25 & 0.0594 & 0.9969 \\
DS18 & 30K & 300 & 20 & Double Gaussian & 250 & 250 & 500 & 0.0738 & 0.9997 \\
\bottomrule
\end{tabular}%
}
\vspace{-.5cm}
\end{table}

\begin{table*}[t]
\scriptsize
\centering
\captionsetup{justification=centering}
\setlength{\tabcolsep}{3pt}
\caption{Experiment \ref{sec:drift_robustness} -- Drift Robustness
\vspace{-.1cm}
\begin{minipage}[t]{\textwidth}
\centering
\tiny
\textbf{Drift Types:} A = Abrupt, I = Incremental, G = Gradual. Synthetic drifts are injected by altering function parameters: abrupt (sudden change), incremental (smooth shift from one concept to the next), and gradual (transition via alternating between two concepts before stabilizing).
\textbf{Concept Loc.:} Points per concept (I. Drift recurs every $\tilde{N}$ points);
\textbf{Y KS P-Value:} Kolmogorov–Smirnov test p-value;
\textbf{Y JS Div:} Jensen–Shannon Divergence;
\textbf{Y Wasserstein Dist:} Normalized Wasserstein distance for $y$;
\textbf{X Any KS Sig:} True if any feature shows significant KS drift ($p < 0.05$);
\textbf{X Min KS P-Value:} Minimum KS p-value across features;
\textbf{X Avg JS Div:} Average JS divergence across features; \quad \textbf{$c_{i \rightarrow j}$:} Any two consecutive concepts. \\
\textbf{Conf. Parameters:} Initial Kernel = RBF; Uncertainty Threshold = 0.001; Decay $\gamma$ = 0.99; KPI = R$^2$; $\rho$ = 0.006; Safe Area Threshold ($\zeta$)= 0.005. 
(DS020, DS022, DS024): Initial = 50, Increment = 20, Max Inducing = 100; \quad
(DS021, DS023, DS025): Initial = 100, Increment = 100, Max Inducing = 200. \textbf{Evaluation} uses an 80/20 temporal split.
\end{minipage}
}

\label{tab:drift-robustness-datasets}

\resizebox{\textwidth}{!}{%
\begin{tabular}{cccccccccccccc}
\toprule
\makecell{Data\\sets} & \makecell{Drift\\Type}&\makecell{Data\\Points} & \makecell{Dimen-\\sions} & Noise & \makecell{Concept\\Loc.} &\makecell{Y KS \\P-Value}
& \makecell{Y JS\\Div} & \makecell{Y Wasserstein\\Dist} & \makecell{X Any\\KS Sig } & \makecell{X Min\\KS P-Value} & \makecell{X Avg\\JS Div}
& MSE & R$^2$ \\

\midrule
DS020 & A & 3k & 3 & 0.1 & 1.5k & 0.0 & 0.8036 & 0.4451 & T & 0.0 & 0.8325 & 0.0012 &0.9994\\
DS021 & A & 10k & 50 & 0.2 & 5k & 0.0 & 0.8326 & 0.5848 & T &  0.0 & 0.8324 & 0.0080& 0.9182\\
DS022 & I & 2k & 3 & 0.1 & every .5k & $c_{i \rightarrow j}$: 0.0 & $c_{i \rightarrow j}$: 0.6469 & 0.3848 ± 0.0173 & $c_{i \rightarrow j}$: T &  $c_{i \rightarrow j}$: 0.0& 0.6624 ± 0.0053 & 0.0034 & 0.9789\\
DS023 & I & 10k & 50 & 0.2 & every 2.5k & $c_{i \rightarrow j}$: 0.0 & 0.6113 ± 0.0037 & 0.3238 ± 0.0112 & $c_{i \rightarrow j}$: T & $c_{i \rightarrow j}$: 0.0 & 0.6797 ± 0.0007 & 0.0134 & .9209\\
DS024 & G & 1.6k & 5 & 0.1 & \makecell{[c1:500, c2: 150, c3:150,\\ c4:150, c5:150, c6:300]} & $c_{i \rightarrow j}$: 0.0 & $c_{i \rightarrow j}$: 0.8326 & 0.5482 ± 0.0043 & $c_{i \rightarrow j}$: T & $c_{i \rightarrow j}$: 0.0 &  $c_{i \rightarrow j}$: 0.8326&0.0057& 0.972\\
DS025 & G & 5.2k & 50 & 0.2 & \makecell{[c1:2.5k, c2: 0.5k,c3: 0.5k,\\ c4: 0.5k, c5: 0.5k, c6: 2.5k]} & $c_{i \rightarrow j}$: 0.0 & 0.8310 ± 0.0010 & 0.4967 ± 0.0176 & $c_{i \rightarrow j}$: T & $c_{i \rightarrow j}$: 0.0 & $c_{i \rightarrow j}$:  0.8326  & 0.0151&0.9220 \\
\bottomrule
\end{tabular}%
}
\end{table*}
\begin{figure*}[t]
    \centering

    \begin{minipage}{0.32\textwidth}
        \centering
        \includegraphics[width=\linewidth]{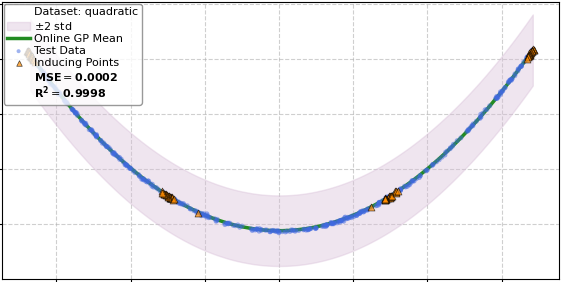}
    \end{minipage}
    \hfill
    \begin{minipage}{0.32\textwidth}
        \centering
        \includegraphics[width=\linewidth]{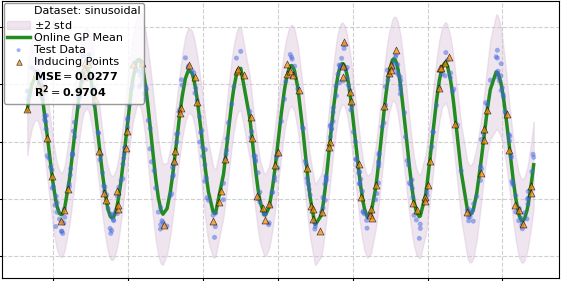}
    \end{minipage}
    \hfill
    \begin{minipage}{0.32\textwidth}
        \centering
        \includegraphics[width=\linewidth]{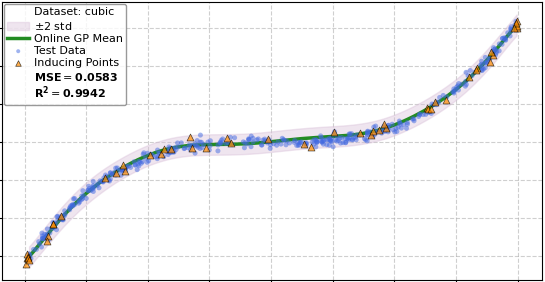}
    \end{minipage}

    \vspace{0.1cm}

    \begin{minipage}{0.32\textwidth}
        \centering
        \includegraphics[width=\linewidth]{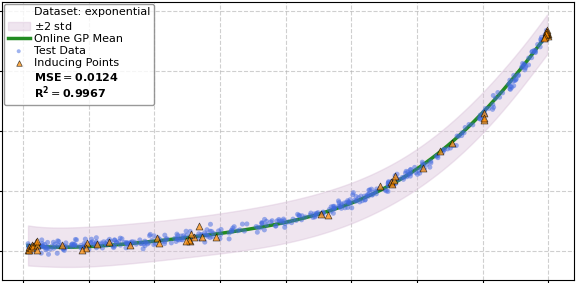}
    \end{minipage}
    \hfill
    \begin{minipage}{0.32\textwidth}
        \centering
        \includegraphics[width=\linewidth]{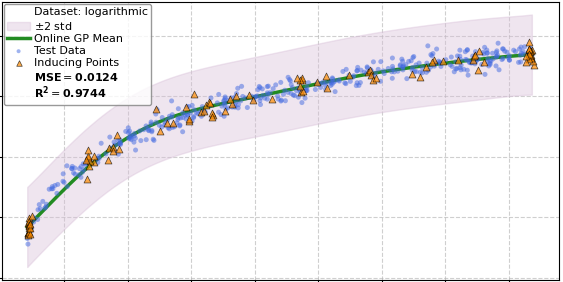}
    \end{minipage}
    \hfill
    \begin{minipage}{0.32\textwidth}
        \centering
        \includegraphics[width=\linewidth]{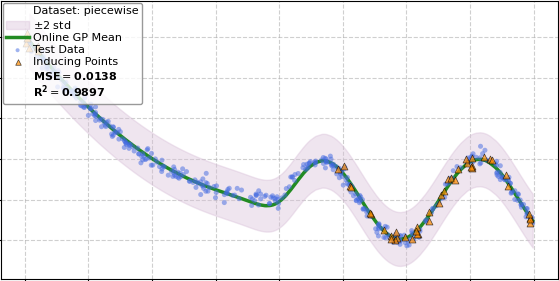}
    \end{minipage}

    \vspace{0.1cm}

    \begin{minipage}{0.32\textwidth}
        \centering
        \includegraphics[width=\linewidth]{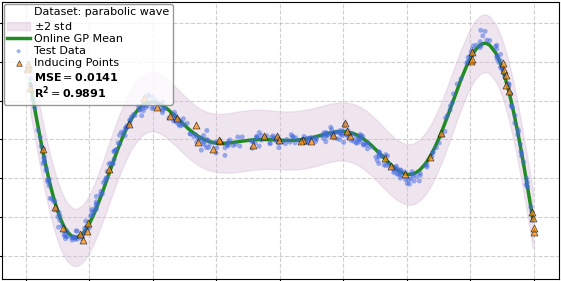}
    \end{minipage}
    \hfill
    \begin{minipage}{0.32\textwidth}
        \centering
        \includegraphics[width=\linewidth]{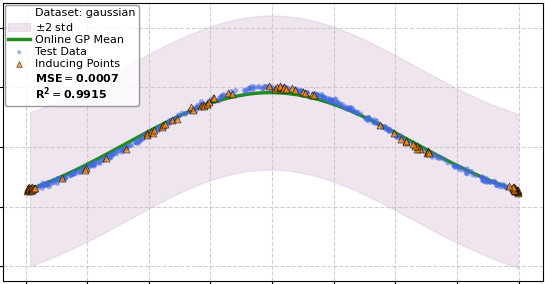}
    \end{minipage}
    \hfill
    \begin{minipage}{0.32\textwidth}
        \centering
        \includegraphics[width=\linewidth]{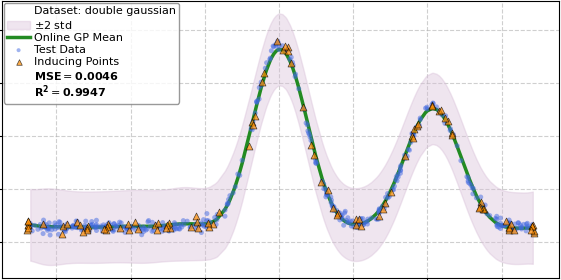}
    \end{minipage}

\caption{
        \scriptsize Experiment ~\ref{sec:2d_visualization}. DAO-GP Prediction and Uncertainty on Multiple Stationary Non-linear Functions.
    }\vspace{-.67ex}
    \scalebox{0.75}{%
        \begin{minipage}{\textwidth}
            \centering
            \tiny
            \textbf{*Reproducibility:} \quad 
            $\bullet$ \textbf{Conf. Parameters:}  
            Initial batch size = 100, Increment size = 10, Decay $\gamma$ = 1 (stationary), Uncertainty threshold = 0.001, Initial kernel = RBF, Max inducing points = 100, KPI = R$^2$, $\rho$ = 0.0002, Safe area threshold = 0.005.
            \\
            $\bullet$ \textbf{Dataset:} Samples = 2000, Features = 2, Noise level = 0.1, Test portion (Temporal Split) = 20\%, Function specified per figure.
        \end{minipage}
    }
    \label{fig:nonlinear_2d_results}
    \vspace{-.5cm}
\end{figure*}
\begin{figure*}[t]
    \centering

    \begin{minipage}{0.32\textwidth}
        \centering
        \includegraphics[width=\linewidth]{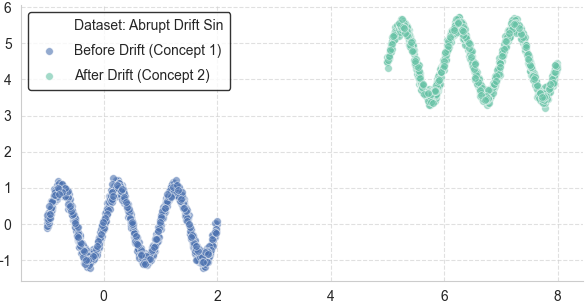}
    \end{minipage}
    \makebox[0pt][c]{\Large$\Rightarrow$}
    \hfill
    \begin{minipage}{0.32\textwidth}
        \centering
        \includegraphics[width=\linewidth]{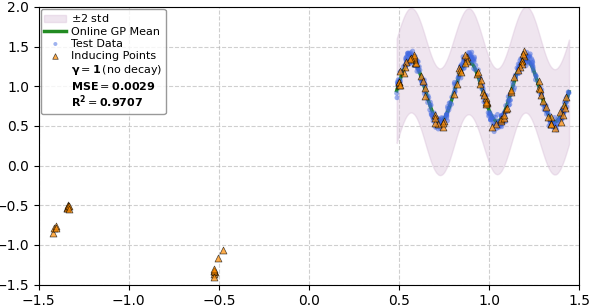}
    \end{minipage}
    \hfill
    \begin{minipage}{0.32\textwidth}
        \centering
        \includegraphics[width=\linewidth]{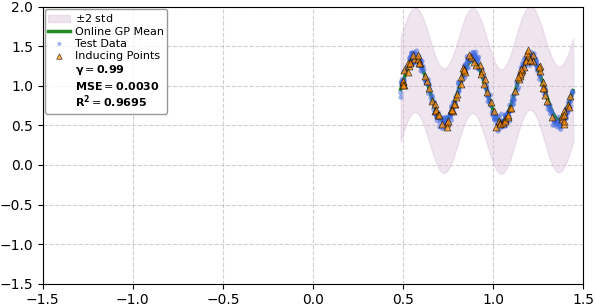}
    \end{minipage}


    \begin{minipage}{0.32\textwidth}
        \centering
        \includegraphics[width=\linewidth]{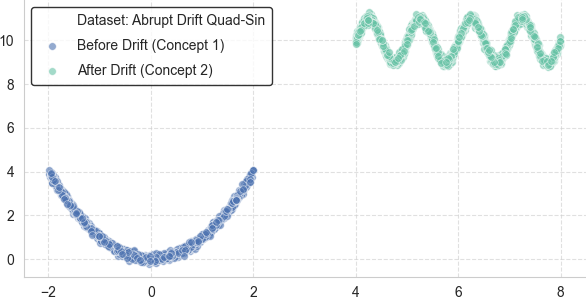}
    \end{minipage}
    \makebox[0pt][c]{\Large$\Rightarrow$}
    \hfill
    \begin{minipage}{0.32\textwidth}
        \centering
        \includegraphics[width=\linewidth]{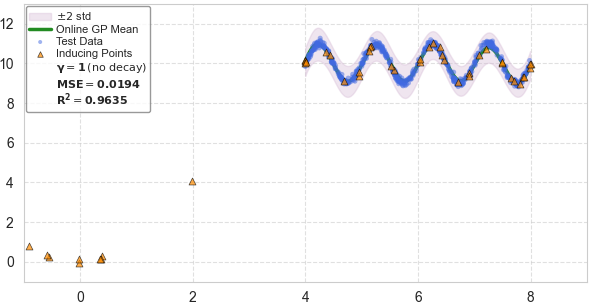}
    \end{minipage}
    \hfill
    \begin{minipage}{0.32\textwidth}
        \centering
        \includegraphics[width=\linewidth]{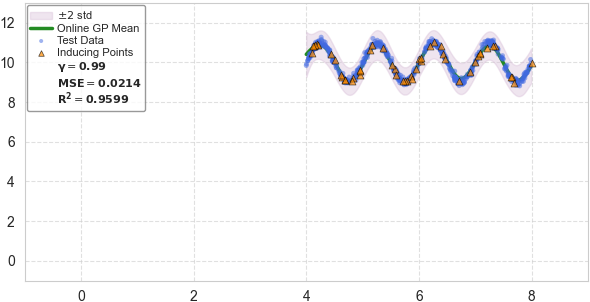}
    \end{minipage}

    \vspace{0.1cm}

    \begin{minipage}{0.32\textwidth}
        \centering
        \includegraphics[width=\linewidth]{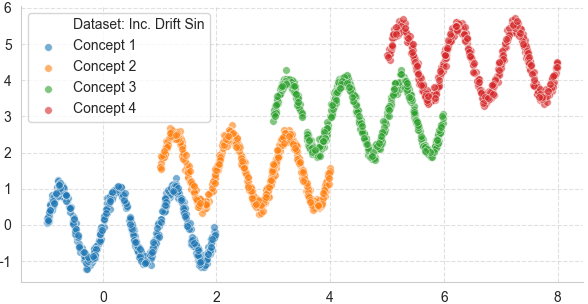}
    \end{minipage}
    \makebox[0pt][c]{\Large$\Rightarrow$}
    \hfill
    \begin{minipage}{0.32\textwidth}
        \centering
        \includegraphics[width=\linewidth]{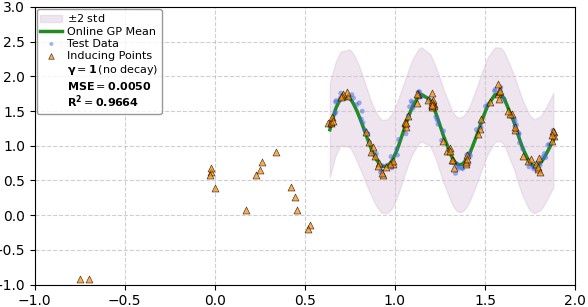}
    \end{minipage}
    \hfill
    \begin{minipage}{0.32\textwidth}
        \centering
        \includegraphics[width=\linewidth]{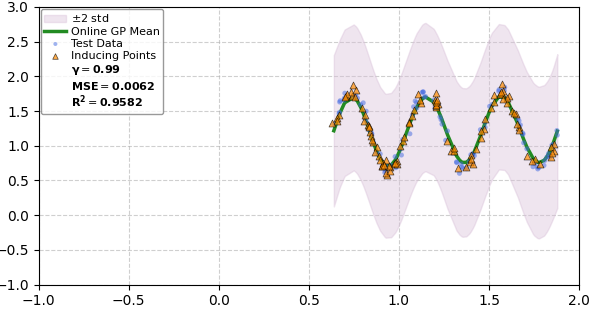}
    \end{minipage}

    \begin{minipage}{0.32\textwidth}
        \centering
        \includegraphics[width=\linewidth]{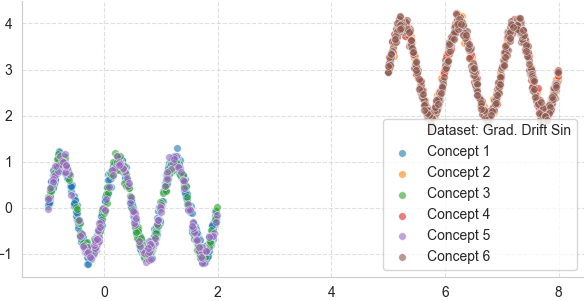}
    \end{minipage}
    \hspace{.1cm}\makebox[0pt][c]{\Large$\Rightarrow$}
    \hfill
    \begin{minipage}{0.32\textwidth}
        \centering
        \includegraphics[width=\linewidth]{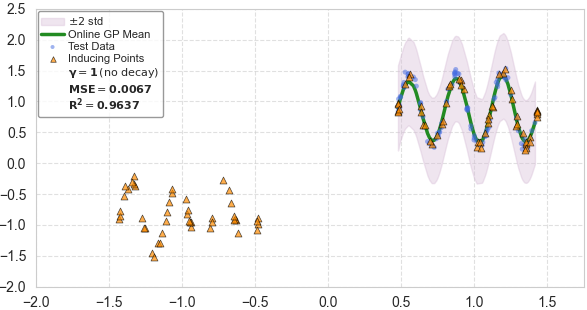}
    \end{minipage}
    \hfill
    \begin{minipage}{0.32\textwidth}
        \centering
        \includegraphics[width=\linewidth]{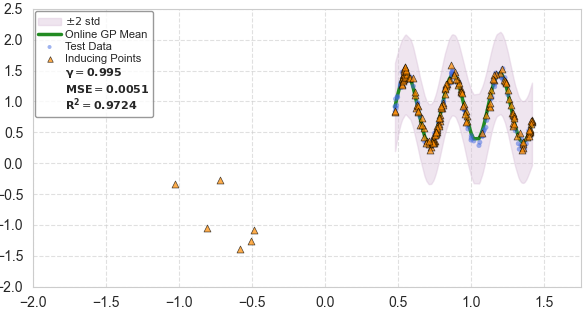}
    \end{minipage}

\caption{
        \scriptsize Experiment ~\ref{sec:2d_drift_decay_visualization}. DAO-GP Prediction and Uncertainty on Multiple Non-Stationary Non-linear Functions.
    }\vspace{-.8ex}
    \scalebox{0.75}{%
        \begin{minipage}{0.99\textwidth}
            \centering
            \tiny
            \textbf{*Reproducibility:} \quad 
            $\bullet$ \textbf{Conf. Parameters:}  
            Initial batch size = 50-100, Increment size = 20, Uncertainty threshold = 0.001, Initial kernel = RBF, Max inducing points = 50-150, KPI = R$^2$, $\rho$ = 0.0002, Safe area threshold = 0.005.
            \\
            $\bullet$ \textbf{Dataset:} Samples = 2000, Features = 2, Noise level = 0.1, Test portion (Temporal Split) = 20\%, Function specified per figure.
        \end{minipage}
    }
    \label{fig:nonlinear_adrift_decay_2d_results}
    \vspace{-.5cm}
\end{figure*}

\begin{figure*}[ht]
    \centering
    \begin{subfigure}{0.475\textwidth}
        \centering
        \includegraphics[height=0.165\textheight]{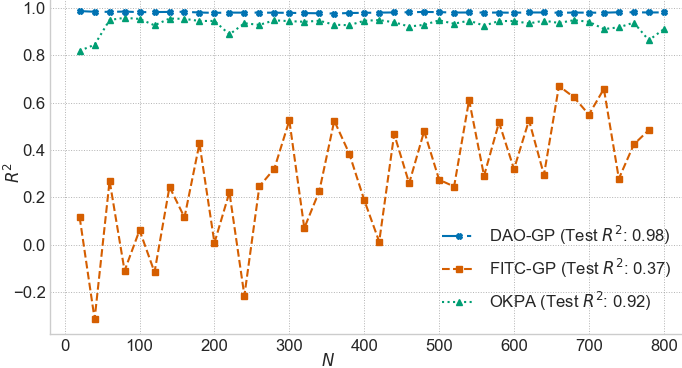}
        \vspace{-.33cm}
        \caption{\scriptsize Models Performance on Stationary DS01}
        \label{fig:sub1}
    \end{subfigure}
\hfill
    \begin{subfigure}{0.475\textwidth}
        \centering
        \includegraphics[height=0.165\textheight]{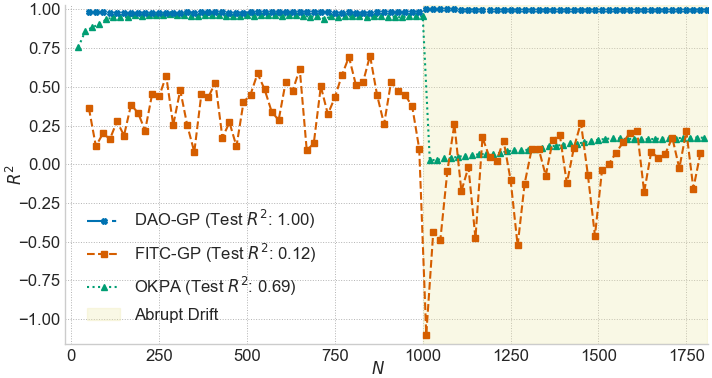}
        \vspace{-.33cm}
        \caption{\scriptsize Models Performance on Abrupt Drift DS021}
        \label{fig:sub2}
    \end{subfigure}
    \begin{subfigure}{0.475\textwidth}
        \centering
        \includegraphics[height=0.165\textheight]{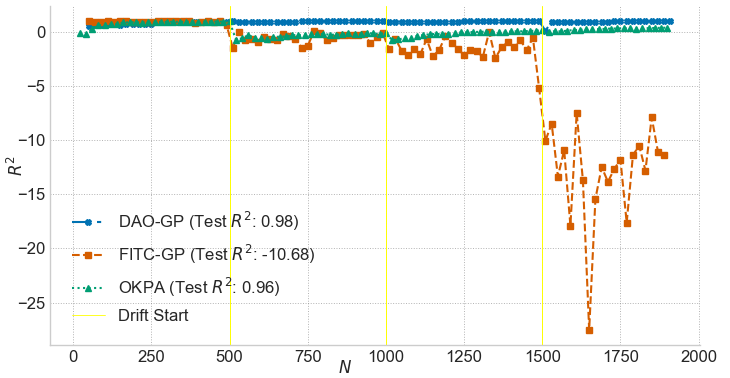}
        \vspace{-.33cm}
        \caption{\scriptsize Models Performance of Incremental Drift DS023}
        \label{fig:sub3}
    \end{subfigure}
    \hfill
    \begin{subfigure}{0.475\textwidth}
        \centering
        \includegraphics[height=0.165\textheight]{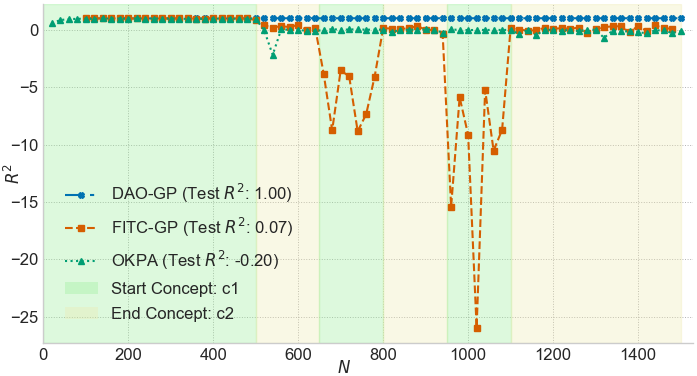}
        \vspace{-.33cm}
        \caption{\scriptsize Models Performance on Gradual Drift DS025}
        \label{fig:sub4}
    \end{subfigure}
    \begin{subfigure}{0.475\textwidth}
        \centering
        \includegraphics[height=0.165\textheight]{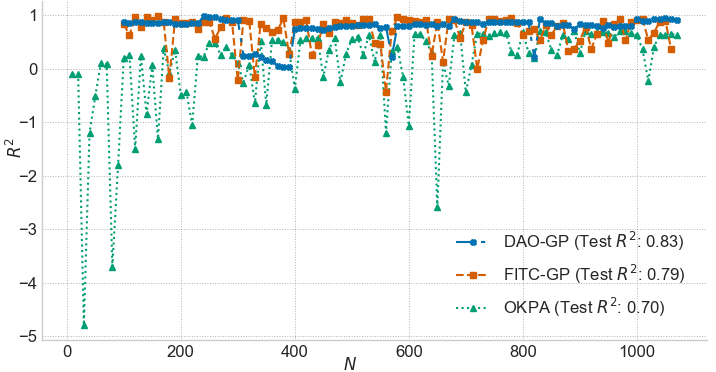}
        \vspace{-.33cm}
        \caption{\scriptsize Models Performance of Medical Insurance Cost Dataset \cite{dataset-medical-insurance}}
        \label{fig:sub3}
    \end{subfigure}
    \hfill
    \begin{subfigure}{0.475\textwidth}
        \centering
        \includegraphics[height=0.165\textheight]{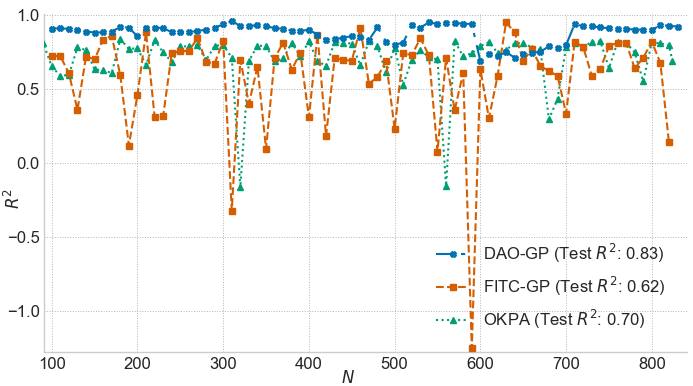}
        \vspace{-.33cm}
        \caption{\scriptsize Models Performance on Concrete Compressive Strength Dataset \cite{dataset-concrete}}
        \label{fig:sub4}
    \end{subfigure}
    \vspace{-.1cm}
\caption{\footnotesize Performance Comparison of DAO-GP, KPA, and FITC-GP Under Varied Streaming Conditions 
}
\vspace{-.35cm}
\label{fig:exp2_incremental}
\end{figure*}

This section presents a comprehensive evaluation of the proposed model across a wide range of experimental categories.  We describe the datasets used, the evaluation metrics employed, and the specific configurations for each experiment. All experiments follow a strictly online setup, processing data sequentially without storage.
Our goal is to demonstrate the effectiveness of DOA-GP and provide a comprehensive analysis of its performance under various conditions.

\textbf{Evaluation Metric:} The coefficient of determination (R$^2$)~\cite{ozer1985correlation} and the Mean Squared Error (MSE)\cite{hyndman2006another} are employed to evaluate the predictive performance of DAO-GP and the baseline models. R$^2$ captures the proportion of variance explained, while MSE penalizes larger errors. Together, they provide a balanced view of accuracy and error magnitude. 

\subsection{\textbf{Stationary Non-Linear Regression: 2D Visualization}}\label{sec:2d_visualization}
This experiment, depicted in Figure \ref{fig:nonlinear_2d_results}, visualizes the model's predictions and associated uncertainty across diverse stationary 2D non-linear regression scenarios. The results consistently demonstrate DAO-GP’s robust predictive performance and well-calibrated uncertainty estimates across a wide range of function types. Notably, inducing points are strategically placed in regions of higher uncertainty, and the model consistently achieves high accuracy, with R$^2$ values exceeding 0.97 even on complex functions.

\subsection{\textbf{Robustness Under Varying Conditions}}\label{sec:robustness-under-varying-conditions}
This experiment aims to evaluate the robustness of DAO-GP under varying data characteristics, including changes in size, dimensionality, noise level, and functional complexity. As shown in Table~\ref{tab:nnr-dataset-full}, DAO-GP consistently achieves high R$^2$ scores and low MSE values across all scenarios, demonstrating its strong adaptability and stability. These results highlight the effectiveness of the model's adaptation and uncertainty-driven update strategy in maintaining reliable performance under diverse conditions.

\subsection{\textbf{Drift Non-Linear Regression: 2D Visualization}}\label{sec:2d_drift_decay_visualization}
This experiment, illustrated in Figure~\ref{fig:nonlinear_adrift_decay_2d_results}, presents the model’s predictions and uncertainty estimates across various non-stationary 2D non-linear regression scenarios. The focus is on evaluating DAO-GP’s adaptability under different types of concept drift (Abrupt, Incremental, and Gradual), highlighting the impact of the decay mechanism. Specifically, the experiment compares the model’s performance with decay ($\gamma < 1$) versus without decay ($\gamma = 1$). The results demonstrate that decay facilitates the retention of relevant, recent information by prioritizing recent data in the inducing set. This enhances adaptability to new patterns while maintaining strong performance, as evidenced by consistently low MSE and high R$^2$ scores.

\subsection{\textbf{Drift Non-Linear Regression: Robustness}}\label{sec:drift_robustness}
This experiment, detailed in Table \ref{tab:drift-robustness-datasets}, incorporates a variety of drift types, including abrupt, incremental, and gradual, each with distinct characteristics. The datasets were synthetically designed to introduce drift in both features (X) and the target variable (Y), enabling comprehensive evaluation. To quantify distributional shifts, we employ the Kolmogorov–Smirnov test \cite{massey1951kolmogorov, press2007numerical}, the Jensen–Shannon divergence \cite{endres2003new, lin2002divergence}, and the Wasserstein distance \cite{villani2008optimal}. Specifically, we assess six indicators: the KS p-value on the target variable (Y KS P-Value), where values below 0.005 signal significant drift; the JS divergence on Y (Y JS Div), ranging from 0 (identical) to 1 (divergent); the Wasserstein distance on Y (Y Wasserstein Dist), normalized to [0,1], where values close to 0 indicate minimal displacement and values approaching 1 indicate substantial shifts in the target distribution; a boolean indicator (X Any KS Sig) denoting whether any feature in X shows KS significance (p $<$ 0.05); the minimum KS p-value across X features (X Min KS P-Value); and the average JS divergence across all features in X (X Avg JS Div), providing a holistic view of feature drift.

Table \ref{tab:drift-robustness-datasets} demonstrates clear evidence of concept drift in all datasets, with KS p-values near zero and high JS divergence values for both inputs and outputs. Significant input distribution shifts are consistently detected (X Any KS Sig = T), especially in abrupt and gradual drift scenarios where JS divergence exceeds 0.83. Incremental drift shows smoother changes with slightly lower divergence. Despite these shifts, the model maintains strong performance, achieving consistently high R² scores (avg.\ $\approx 0.952$) and low MSE, indicating effective drift adaptation across varying drift types and data complexities.

\subsection{\textbf{Online Nonlinear Regression Models: Benchmarking}}\label{sec:benchmark-nonlinear-online-regression-models}

In this experiment, we benchmark our model against two representative online nonlinear regression models: Kernelized Passive-Aggressive Regression (KPA), a parametric approach, and FITC-GP, a non-parametric method. These models were selected from the broader set discussed in the related work section based on the availability of reliable code repositories and their performance on challenging datasets. The benchmarking process leverages synthetic and real-world datasets, encompassing scenarios with stationary distributions and various types of non-stationary drift. To ensure a fair and consistent experimental setup, we adopted the following protocols:
\begin{enumerate}[leftmargin=*, labelsep=0.5em]
    \item For both FITC-GP and KPA models, the kernel function used during the initialization phase of DAO-GP (with the same initial mini-batch) was consistently applied.
    \item Recognizing that KPA lacks inherent kernel hyperparameter optimization, we directly applied the optimized kernel hyperparameters obtained from DAO-GP's initialization to KPA for comparable results.
    \item Configuration parameters (e.g., initial batch size, increment size) are standardized across models for consistency.  
\end{enumerate}

Upon evaluating the models across diverse real-world and synthetic datasets under various drift scenarios, it is evident that both KPA and FITC-GP lack robustness to concept drift. For instance, in Experiment (b), both models experience a pronounced performance drop at the onset of drift. Additionally, FITC-GP exhibits considerable instability across nearly all scenarios. While KPA performs adequately in stationary settings, its lack of an online hyperparameter optimization mechanism renders it highly sensitive to initial hyperparameter selection. We initialized KPA with optimized parameters in all experiments, resulting in strong initial performance; however, its effectiveness deteriorates sharply following any distributional shift as those parameters become outdated. In contrast, DAO-GP consistently demonstrates strong performance and robustness across most scenarios. In Experiment (e), a performance drop is observed between points 300 and 400, suggesting that the current kernel pool still has room for improvement to better capture a wider range of data distributions and underlying relationships.

\section{Conclusion \& Future Work}
DAO-GP sets a new benchmark for online Gaussian Process regression by integrating a comprehensive suite of mechanisms tailored for non-stationary streaming environments. Unlike conventional GP models constrained by static assumptions and manual tuning, DAO-GP is architected as a fully adaptive, tuning-free framework that responds autonomously to evolving data distributions. Its built-in drift detection not only identifies shifts but also quantifies their magnitude, enabling selective and efficient re-optimization of kernel hyperparameters only when necessary. By adopting KPI-based detection over distributional methods, DAO-GP remains effective in high-dimensional streams, aligning with the scalability demands of big data. This on-demand adaptation strategy avoids the inefficiencies of batch-wise retraining, consistent with the theoretical foundations of online learning. The model maintains sparsity through a principled inducing point selection scheme that prioritizes informativeness and recency via decay-aware scoring, thereby ensuring bounded complexity and scalability for long-horizon deployments, making it well suited for large-scale data streams. By integrating exponential decay into kernel computation, DAO-GP down-weights past observations, enabling fast adaptation to new patterns while reducing memory and computational cost. Moreover, its dynamic kernel selection from a diverse pool equips it to capture varying degrees of nonlinearity and periodicity across different phases of the data stream. Collectively, these innovations establish DAO-GP as a state-of-the-art, resource-efficient, and drift-resilient solution for real-time non-linear regression under concept drift. Future work includes sensitivity analysis of DAO-GP’s configuration parameters (e.g., KPI, inducing points) and exploring its MLaaS deployment for adaptive online regression with minimal user intervention.

\bibliographystyle{IEEEtran}
\bibliography{references}

\end{document}